\documentclass[runningheads]{llncs}
\usepackage{amsmath}
\usepackage{graphicx}
\usepackage{CJKutf8}
\newcommand{\spara}[1]{\bigskip\noindent{\bf #1}}

\begin{document}
\title{AsdKB: A Chinese Knowledge Base for the Early Screening and Diagnosis of Autism Spectrum Disorder}
\titlerunning{AsdKB: A Chinese Knowledge Base for Autism Spectrum Disorder}
%
\author{Tianxing Wu\inst{1,2}\and Xudong Cao\inst{1}\and
Yipeng Zhu\inst{1}\and Feiyue Wu\inst{1}\and Tianling Gong\inst{1}\and Yuxiang Wang\inst{3}\and Shenqi Jing\inst{4,5}}
\authorrunning{T. Wu et al.}
%
\institute{Southeast University, China\\
\email{\{tianxingwu, xudongcao, yipengzhu, wufeiyue, gtl2019\}@seu.edu.cn}\\
\and
Key Laboratory of New Generation Artificial Intelligence Technology and Its Interdisciplinary Applications (Southeast University), Ministry of Education, China
\and
Hangzhou Dianzi University, China\\
\email{lsswyx@hdu.edu.cn}
\and
The First Affiliated Hospital of Nanjing Medical University, China
\and
Nanjing Medical University, China\\
\email{jingshenqi@jsph.org.cn}}

\maketitle              
\begin{abstract}
To easily obtain the knowledge about autism spectrum disorder and help its early screening and diagnosis, we create AsdKB, a Chinese knowledge base on autism spectrum disorder. The knowledge base is built on top of various sources, including 1) the disease knowledge from SNOMED CT and ICD-10 clinical descriptions on mental and behavioural disorders, 2) the diagnostic knowledge from DSM-5 and different screening tools recommended by social organizations and medical institutes, and 3) the expert knowledge on professional physicians and hospitals from the Web. AsdKB contains both ontological and factual knowledge, and is accessible as Linked Data at \textbf{\url{https://w3id.org/asdkb/}}. The potential applications of AsdKB are question answering, auxiliary diagnosis, and expert recommendation, and we illustrate them with a prototype which can be accessed at \textbf{\url{http://asdkb.org.cn/}}.

\end{abstract}

\section{Introduction}
Autism spectrum disorder (ASD) is a kind of neurodevelopmental disability which begins before the age of 3 years and can last throughout a person’s whole life. People with ASD have problems in social communication and interaction, and may have stereotypic or repetitive behaviors (or interests). According to the most recent statistics~\cite{maenner2021prevalence} published by the Centers for Disease Control and Prevention (CDC), about 1 in 36 children aged 8 years has been identified with ASD, and this proportion is quite high. However, there is no quantitative medical test to diagnose such a disorder, and professional physicians only use screening tools and look at the behaviors for some time to make a diagnosis. In this way, many children cannot receive a final diagnosis until much older, which causes the children with ASD might not get the early help they need. In China, the situation on screening and diagnosing the children with ASD maybe much worse compared with western developed countries. The 2020 China rehabilitation report of children developmental disorder\footnote{\url{http://pkucarenjk.com/news-family/2303.html}} points out that the ASD incidence in China is around 1\% and the number of ASD children is more than three million, but the number of professional physicians who can diagnose ASD is only about 500, let alone the number of board certified behavior analysts. This does hinder the timely diagnosis on ASD, which inspires us to think about if we can apply artificial intelligence techniques to solve the early screening and diagnosis of ASD. The key problem is how to extract and integrate ASD relevant knowledge from heterogeneous sources to support upper-level intelligent applications.
\begin{figure}[t]
  \centering
  \includegraphics[width=1\textwidth]{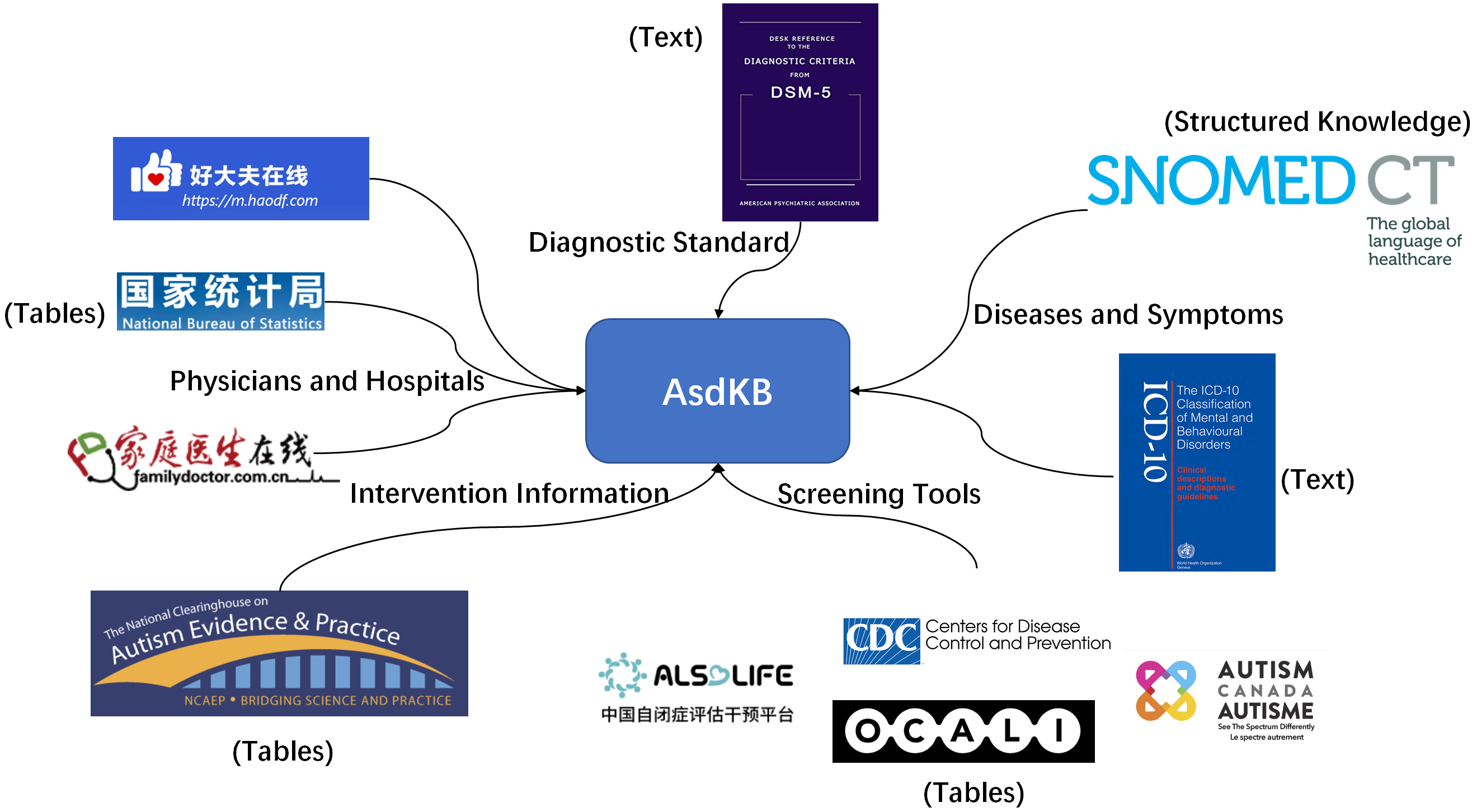}
\caption{The data sources for building AsdKB.}
\label{fig:sources}
\end{figure}

To solve this problem, we build AsdKB, a Chinese knowledge base for the
early screening and diagnosis of ASD, from various sources (see Figure~\ref{fig:sources}), such as SNOMED CT~\cite{donnelly2006snomed} (a large collection of medical terms), ICD-10\footnote{\url{https://en.wikipedia.org/wiki/ICD-10}} (the $10$th revision of the classification system of diseases published by WHO) clinical descriptions on mental and behavioural disorders~\cite{world1992icd}, DSM-5~\cite{american2013diagnostic} (the $5$th edition of diagnostic and statistical manual of mental disorders), the screening tools recommended by CDC and so on. Specifically, we first build an ontology covering important concepts about the screening and diagnosis of ASD from DSM-5, ICD-10 clinical descriptions on mental and behavioural disorders, SNOMED CT, CDC materials, and other Web sources. Using this ontology as the schema, we then extract and integrate factual knowledge on diseases, diagnosis, experts, and others. Besides, we use and develop Web crawler and natural language processing (NLP) tools for data extraction, keyword extraction, knowledge extraction, machine translation, and etc., over various formats of data, including text, tables, and structured knowledge. All classes, properties, and instances in AsdKB are identified by permanent dereferenceable URIs in w3id\footnote{\url{https://w3id.org/asdkb/}}. All data are available as RDF dump files on Zenodo\footnote{\url{https://zenodo.org/record/8199698}}, and the basic information of the AsdKB project can be accessed at Github\footnote{\url{https://github.com/SilenceSnake/ASDKB}}. All the resources are published under CC BY-SA 4.0. The main contributions of this paper are summarized as follows:
\begin{itemize}
    \item We first build a Chinese knowledge base for the early screening and diagnosis of ASD, i.e., AsdKB, which contains both ontological and factual knowledge, and publish it following Linked Data best practices.
    \item We present a prototype system on question answering, auxiliary diagnosis, and expert recommendation with AsdKB, and discuss how to support the early screening and diagnosis of ASD with this system.
\end{itemize}

The rest of this paper is organized as follows. Section~\ref{ob} introduces the process of ontology building. Section~\ref{fke} describes the extraction of factual knowledge. Section~\ref{aoa} presents the potential applications of AsdKB. Section~\ref{rw} outlines related work, and we conclude in the last section.
\section{Ontology Building}\label{ob}
\begin{figure}[t]
  \centering  \includegraphics[width=1\textwidth]{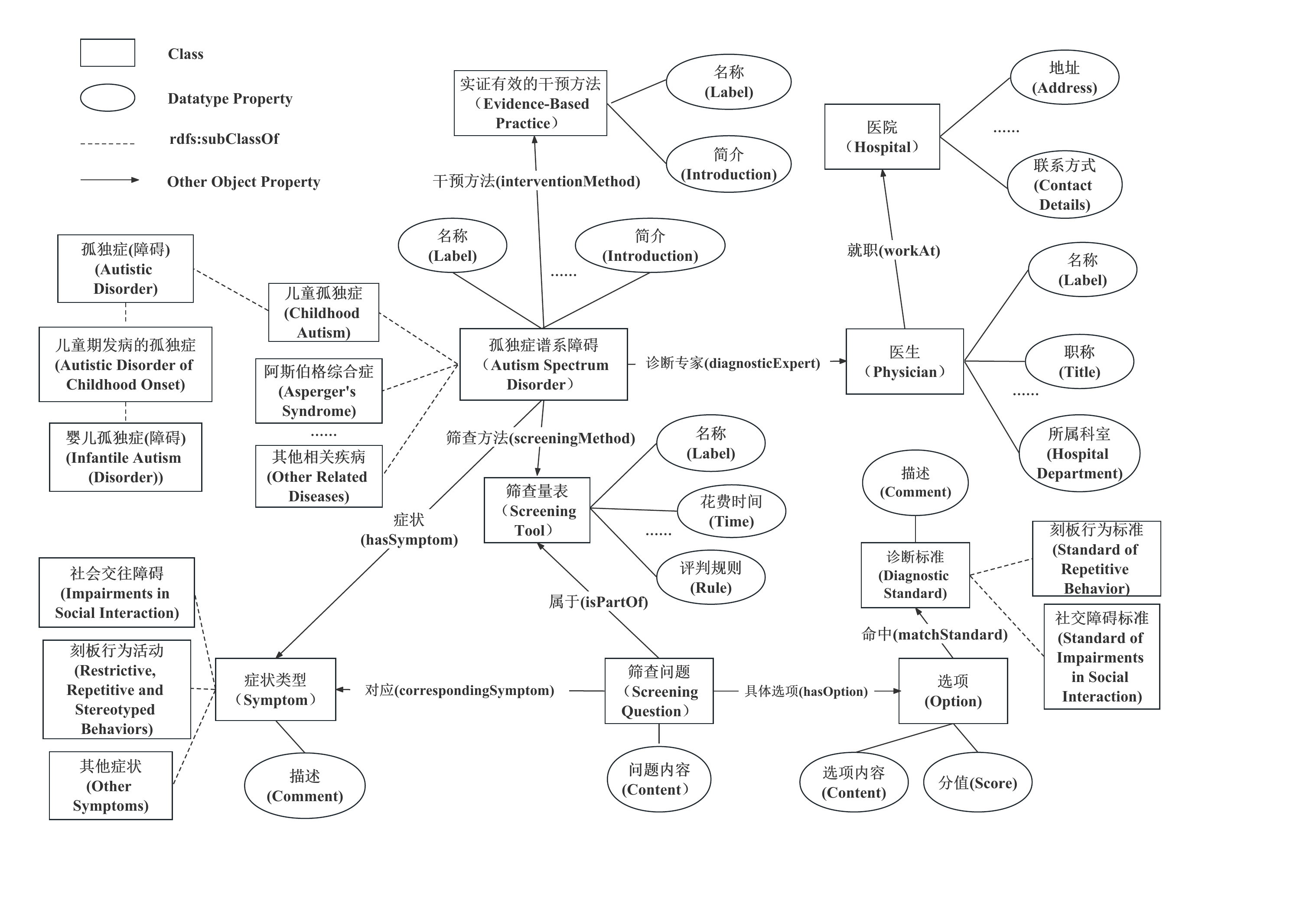}
\caption{A part of the AsdKB ontology.}
\label{fig:ontology}
\end{figure}

This section introduces the process of building the AsdKB ontology as the schema which is used to guide extracting and integrating factual knowledge from various sources. We follow Ontology Development 101~\cite{noy2001ontology} to build the ontology (Figure~\ref{fig:ontology} shows a part of it) as follows.

\spara{Step 1: Determine the domain and scope of the ontology.} AsdKB is expected to cover the ASD relevant knowledge on the early screening and diagnosis, so the ontology needs to cover important concepts in widely recognized materials about the screening and diagnosis of ASD. Here, we select relevant materials from CDC, DSM-5, ICD-10, SNOMED-CT, and other Web sources.

\spara{Step 2: Consider reusing existing ontologies.}  In this part, we reuse the standard RDF, RDFS, and OWL vocabularies, including \texttt{rdf:type} linking from instances to classes, \texttt{rdfs:label} recording the Chinese (or English) labels of classes and properties, \texttt{rdfs:comment} providing textual descriptions to clarify meanings of classes, \texttt{rdfs:subClassOf} describing the class hierarchy, equivalent classes are linked by \texttt{owl:equivalentClass} from the AsdKB ontology to other ontologies, and \texttt{rdfs:domain} and \texttt{rdfs:range} specifying the resources and values of a property are instances of one or more classes, respectively.

\spara{Step 3: Enumerate important terms in the ontology.} We read the ASD materials from CDC, DSM-5, ICD-10, SNOMED CT and other Web sources mentioned in the first step, to manually identify a list of important concept-level terms. For example, important symptom concepts in disease knowledge include ``Impairments in Social Interaction'' and  `` Restrictive, Repetitive and Stereotyped Behaviors''. Important concepts in expert knowledge include ``Physician'' and  ``Hospital''. Besides, ``Screening Tool'' and ``Diagnostic Standard'' are related to screening and diagnosis. 

\spara{Step 4: Define the classes and the class hierarchy.} Based on the previous identified important terms, we start to create disease classes (e.g., ``Autism Spectrum Disorder'' and ``Asperger's Syndrome''), diagnosis classes (e.g., ``Screening Tool'' and ``Screening Question''), expert classes (e.g., ``Physician'' and ``Hospital''), and others. For the class hierarchy, we consider the hierarchies within disease classes, symptom classes, and diagnosis classes, respectively. For example, as shown in Figure~\ref{fig:ontology}, we have `` Asperger's Syndrome \texttt{rdfs:subClassOf} Autism Spectrum Disorder'' and ``Standard of Social Interaction \texttt{rdfs:subClassOf} Diagnostic Standard''. Specifically, we have created a class ``Screening Question'' in the diagnosis classes to facilitate the exploration of the association between instances of ``Screening Question'' and ``Diagnostic Standard''.

\spara{Step 5: Define the properties of classes.} After selecting classes from the list of terms, we start to attach properties to classes using \texttt{rdfs:domain}. We distinguish datatype properties and object properties. For example, for the class ``Physician'', we have the object property \texttt{workAt} and datatype properties \texttt{Name}, \texttt{Title}, \texttt{Specialty}, \texttt{Hospital Department}, and etc.

\spara{Step 6: Define the facets of the properties.} We specify the value type of each property by defining \texttt{rdfs:range}. The range of a datatype property is an XML Schema datatype. For example, the ranges of properties \texttt{Address} (attached to the class ``Hospital'') and \texttt{Score} (attached to the class ``Option'') are \texttt{xsd:string} and \texttt{xsd:float}, respectively. Besides, the range of an object property is a class. For example, the range of \texttt{hasSymptom} is the ``Symptom''.

\spara{Step 7: Create instances.} We do not define instances in the ontology but only use it as the schema of AsdKB. The creation of instances belongs to factual knowledge extraction, and it will be described in Section~\ref{fke}.

\spara{Statistics about the AsdKB ontology.}
The built ontology is currently online: \textbf{\url{https://w3id.org/asdkb/ontology/}}. It contains 32 classes, 25 datatype properties, and 16 object properties. The maximum depth of a class in the class hierarchy is 4. Note that we apply google translate\footnote{\url{https://translate.google.com/}} to translate the English labels of all elements in the ontology into Chinese ones, and also perform careful manual proofreading and correction.

\spara{Mapping to other ontologies.}
To facilitate schema knowledge sharing across different ontologies, We map our AsdKB ontology to the Unified Medical Language System~\cite{bodenreider2004unified} (UMLS) and the Autism DSM-ADI-R (ADAR) ontology~\cite{mugzach2015ontology}. UMLS is the largest integrated biomedical ontology covering the vocabularies from around 200 sources including SNOMED CT, DSM-5, FMA~\cite{rosse2008foundational}, and etc. ADAR is built from Autism Diagnostic Interview-Revised~\cite{lord1994autism} (ADI-R), which is a structured interview used for autism diagnosis. ADAR focuses on autism symptom classification, and it constructs many fine-grained symptom classes, e.g., ``First walked unaided'' and ``Daily spontaneous and meaningful speech'', which are taken as instances in AsdKB. This is why we do not directly re-use ADAR in AsdKB.

\begin{figure}[t]
  \centering
  \includegraphics[width=1\textwidth]{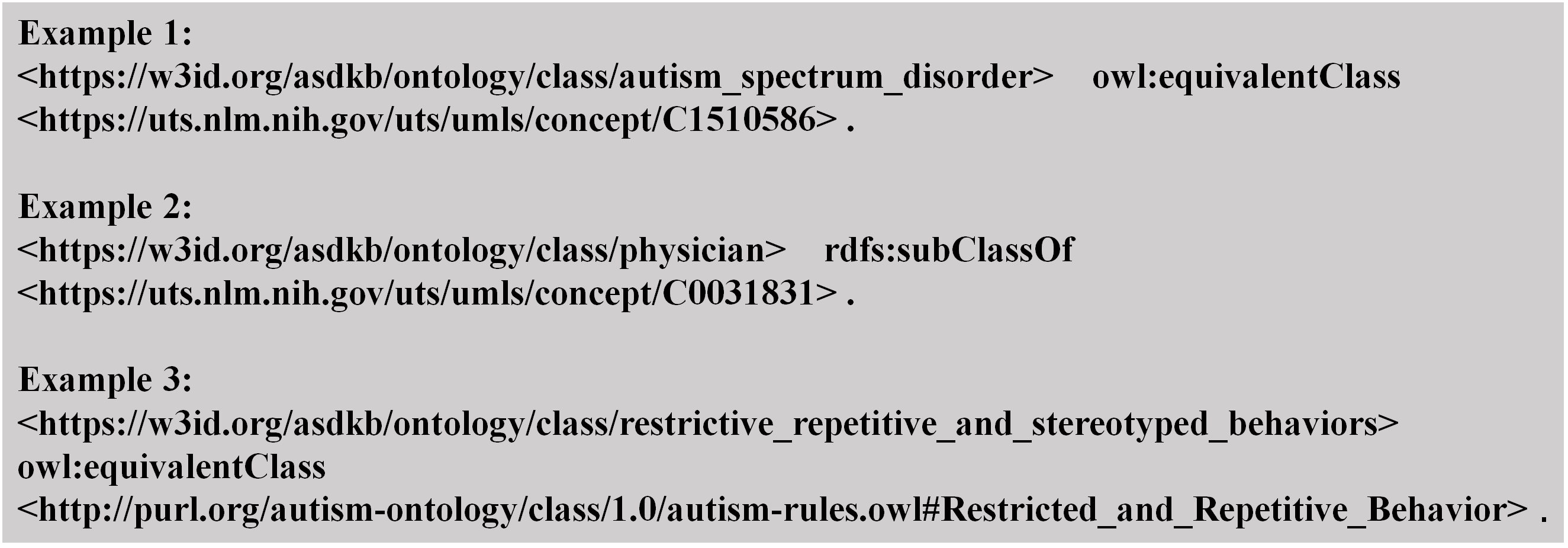}
\caption{Examples of mapping the AsdKB ontology to UMLS and ADAR.}
\label{fig:mapExp}
\end{figure}

Since the disease classes in the AsdKB ontology are extracted from SNOMED CT, which is also a part of UMLS, such classes are naturally linked to UMLS (see the Example 1 in Figure~\ref{fig:mapExp}). For the rest eighteen classes in AsdKB, we submit each of their labels to UMLS Metathesaurus Browser\footnote{\url{https://uts.nlm.nih.gov/uts/umls/home}} and manually make comparisons between the returned classes and submitted ones to decide whether there exist \texttt{owl:equivalentClass} or \texttt{rdfs:subClassOf} relations (the Example 2 in Figure~\ref{fig:mapExp} gives a mapping result). Besides, we apply AgreementMakerLight~\cite{faria2013agreementmakerlight} to mapping the AsdKB ontology to ADAR, and the Example 3 in Figure~\ref{fig:mapExp} also shows a mapping result. 

\section{Factual Knowledge Extraction}\label{fke}
This section presents the extraction of factual knowledge of ASD. Due to limited spaces, we do not explain every detail but focus on the main content.

\subsection{Disease Knowledge} 
For disease knowledge, we need to extract the factual knowledge about disease and symptom instances according to the AsdKB ontology. Disease instances (e.g., ``Atypical Rett syndrome'') are derived from SNOMED CT, and they are actually the leaf nodes in the disease taxonomy in SNOMED CT. For each disease instance, we extract the values of the properties: \texttt{Label} (instance name), \texttt{SCTID} (the term ID in SNOMED CT), \texttt{ICD-10 code} (the corresponding ICD-10 category), and \texttt{Synonym} from SNOMED CT, respectively. We also manually design templates (i.e., regular expressions) to extract the values of properties \texttt{Introduction} (a brief description of the given disease instance), \texttt{Patient Groups} (e.g., ``children'' or ``female children''), and \texttt{Pathogeny} (e.g., ``genetic and environmental factors'') from ICD-10 clinical descriptions on mental and behavioural disorders~\cite{world1992icd}, respectively. Besides, for the values of properties \texttt{Label}, \texttt{Synonym}, \texttt{Introduction}, \texttt{Patient Groups}, and \texttt{Pathogeny}, we obtain the corresponding Chinese versions by Google Translate and manual proofreading. We collect $49$ disease instances relevant to ASD in total, and their corresponding property information.

\begin{figure}[htbp]
  \centering
  \includegraphics[width=1\textwidth]{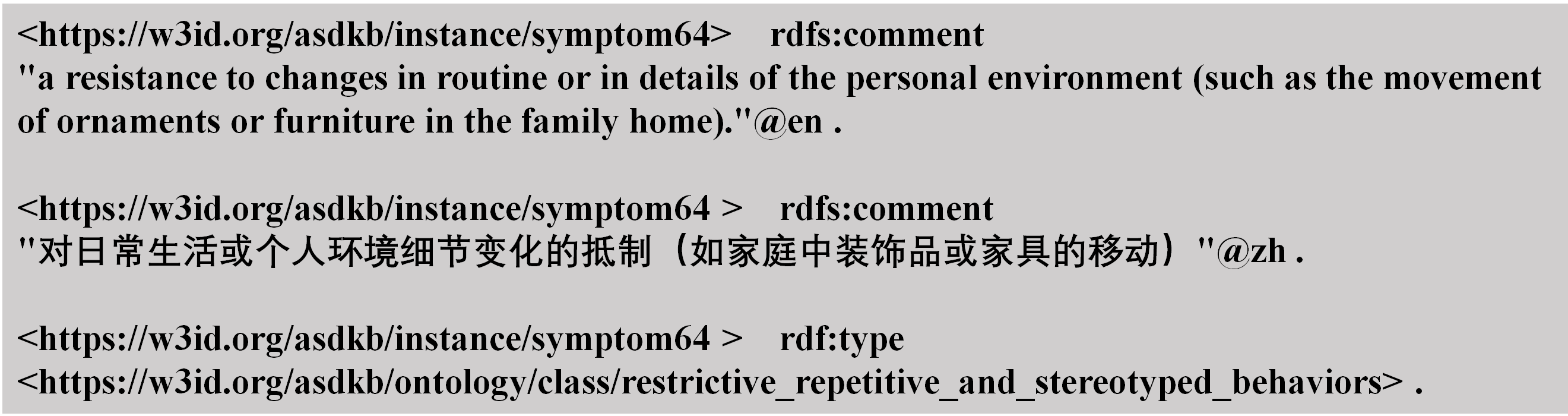}
\caption{An example of the triples containing a symptom instance.}
\label{fig:siExp}
\end{figure}

Symptom instances are also extracted from ICD-10 clinical descriptions on mental and behavioural disorders. We model the symptom instance extraction as the task of sequence labeling. We first take each paragraph as a document, and apply Term Frequency-Inverse Document Frequency~\cite{leskovec2020mining} (TF-IDF) to identify keywords. Based on this, we then label a small amount of symptom instances in the corpus to train an extraction model. Here, we use BioBERT~\cite{lee2020biobert}, a pre-trained biomedical language representation model for biomedical text mining, to encode each word as an embedding. Afterwards, we utilize BiLSTM~\cite{graves2005framewise} to capture textual context features for each word. Finally, we apply conditional random fields~\cite{lafferty2001conditional} to finishing sequence labeling, which naturally classifies symptom instances to the pre-defined symptom classes, i.e., ``Impairments in Social Interaction'', `` Restrictive, Repetitive and Stereotyped Behaviors'', and ``Other Symptoms''. High-quality results of sequence labeling obtained by the trained model will be added to the labeled data to train a new model. We repeat this process until the maximum number of iterations is reached. Google Translate is also used to get the Chinese description of each symptom instance. Figure~\ref{fig:siExp} shows the triples of the symptom \texttt{<\url{https://w3id.org/asdkb/instance/symptom64}>}. Finally, We collect 65 symptom instances in total.

\subsection{Diagnostic Knowledge}\label{subsec:dk}
For diagnostic knowledge, we extract the factual knowledge on the instances of diagnostic standards, screening tools, screening questions, and the corresponding options. Instances of diagnostic standards are acquired from the Chinese edition\footnote{\url{https://www.worldcat.org/zh-cn/title/1052824369}} of DSM-5~\cite{american2013diagnostic}, so we only have Chinese descriptions for the instances of diagnostic standards. We follow a similar process used for extracting symptom instances, and only replace the pre-trained model BioBERT with a more general model BERT~\cite{kenton2019bert} because BioBERT does not support Chinese but BERT does. Different from symptom instances, instances of diagnostic standards do not refer to specific behaviors or activities of the people with ASD, they actually present textual summarizations for specific classes of diagnostic standards (i.e., ``Standard of Repetitive Behavior'' and ``Standard of Impairments in Social Interaction''). For example, an instance of diagnostic standards in AsdKB is expressed as ``\begin{CJK*}{UTF8}{gbsn}
眼神接触、手势、面部表情、身体定位或言语语调等方面的缺乏、减少或不合规的使用
\end{CJK*} (abnormalities in eye contact and body language or deficits in understanding and use of gestures)'', which corresponds to multiple symptom instances, e.g., ``\begin{CJK*}{UTF8}{gbsn}
不会进行对视\end{CJK*} (without eye contact)'' and ``\begin{CJK*}{UTF8}{gbsn}很少微笑\end{CJK*} (rarely smile)''. We collect 43 instances of diagnostic standards in total.

Regarding instances of screening tools, screening questions, and options, we extract them and the corresponding property values from the websites of social organizations and medical institutes, including CDC\footnote{\url{https://www.cdc.gov/ncbddd/autism/hcp-screening.html\#Tools}}, ALSOLIFE\footnote{\url{https://www.alsolife.com/autism/screen/}} (China ASD Evaluation and Intervention Platform), Autism Canada\footnote{\url{https://autismcanada.org/autism-explained/screening-tools/}}, and OCALI\footnote{\url{https://www.ocali.org/project/assessment\_measures}} (The Ohio Center for Autism and Low Incidence). Instances of screening tools in AsdKB are actually screening scales, which have the properties \texttt{Introduction} (basic information and instructions), \texttt{Author}, \texttt{User} (the one filling in the scale, e.g., parents or teacher), \texttt{Age} (applicable ages of screening targets), \texttt{Time} (the time it takes to fill in the scale), \texttt{Rule} (screening principles and details), and \texttt{Screening Boundary} (the score of screening boundary after finishing the scale). After careful selection, we extract twenty instances of screening tools, containing fifteen English screening scales and five Chinese ones, such as ABC~\cite{krug1980behavior}, CARS2~\cite{schopler2010childhood}, and M-CHAT~\cite{wright2014modified}. Google Translate is used here to translate English scales into Chinese ones, and manual proofreading is also conducted.

\begin{figure}[htbp]
  \centering  
  \includegraphics[width=1\textwidth]{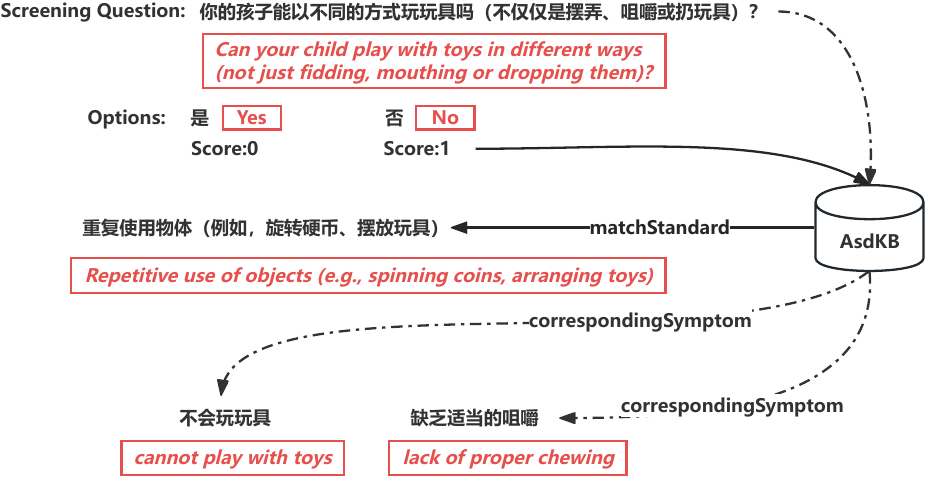}
\caption{An example to show the benefits of \texttt{correspondingSymptom} and \texttt{matchStandard} relationships.}
\label{fig:dkEXP}
\end{figure}

For instances of screening questions and options, they can be directly obtained from screening scales through table extraction. Besides keeping their textual content as the property, we also establish \texttt{correspondingSymptom} relationships between instances of screening questions and symptom instances, and \texttt{matchStandard} relationships between option instances and instances of diagnostic standards. These two kinds of relationships (i.e., object properties) benefit to the interpretability of screening results. For example, as shown in Figure~\ref{fig:dkEXP}, AsdKB can tell users that the current question investigates what specific symptoms are and whether the current option matches some diagnostic standard or not, in order to help users better understand screening questions and provide explanations to screening results. To identify \texttt{correspondingSymptom} relationships, we first use FNLP~\cite{qiu2013fudannlp} to perform Chinese word segmentation on the instances of screening questions and symptom instances. After removing stopwords, we then compare string similarities between two word sequences to decide whether the \texttt{correspondingSymptom} relationship exists. The method of extracting \texttt{matchStandard} relationships is similar to that of \texttt{correspondingSymptom} relationships, and the only difference is to additionally consider the property \texttt{Score} of each option. If an option has the highest score or lowest score, it means the result of the current screening question is abnormal or normal (it ups to the design of screening scales), and abnormal results could help identify \texttt{matchStandard} relationships.

\subsection{Expert Knowledge}\label{subsec:ek}
For expert knowledge, we extract factual knowledge of the instances of professional physicians, and the hospitals they work at, from the Web. We select two famous Chinese healthcare websites 
The Good Doctor\footnote{\url{https://www.haodf.com/}} and Family Doctor\footnote{\url{https://www.familydoctor.com.cn/}} as the data sources to extraction, so the string values of some datatype properties are only presented in Chinese. We first submit the following keywords ``\begin{CJK*}{UTF8}{gbsn}孤独症, 自闭症, 孤独症谱系障碍
\end{CJK*}  (ASD)'',  ``\begin{CJK*}{UTF8}{gbsn}广泛性发育障碍
\end{CJK*} (pervasive developmental disorders)'', ``\begin{CJK*}{UTF8}{gbsn}儿童孤独症
\end{CJK*} (childhood autism)'', and ``\begin{CJK*}{UTF8}{gbsn}阿斯伯格综合症
\end{CJK*} (Asperger's syndrome)'', to the search engines of the selected websites, which locates the Web pages of professional physicians on ASD. Faced with the structures like infobox tables in Wikipedia, we then extract physician instances and the values of properties \texttt{Name}, \texttt{Title} (e.g., ``\begin{CJK*}{UTF8}{gbsn}主任医师\end{CJK*}  (chief physician)'' and ``\begin{CJK*}{UTF8}{gbsn}主治医师\end{CJK*}  (attending physician)''), \texttt{Specialty} (e.g., ``\begin{CJK*}{UTF8}{gbsn}各类儿童精神障碍\end{CJK*} (various types of mental disorders in childhood)''), \texttt{Hospital Department} (e.g., ``\begin{CJK*}{UTF8}{gbsn}儿童保健科\end{CJK*} (child healthcare department)'' and ``\begin{CJK*}{UTF8}{gbsn}精神科\end{CJK*} (psychiatry department)''), and \texttt{workAt} (i.e., hospital instances). We collect 499 physician instances in total.

According to the values of the property \texttt{workAt}, we locate the Web pages of hospital instances. Similar to the extraction on physician instances and the corresponding property information, we extract the hospital instances and the values of properties \texttt{Name}, \texttt{Address}, \texttt{Contact Details}, and \texttt{Hospital Level} (e.g., ``\begin{CJK*}{UTF8}{gbsn}三甲医院\end{CJK*} (Grade-A tertiary hospital)''). We collect 270 hospital instances in total.

Since physician and hospital instances are extracted from different sources, we perform instance matching using heuristics. Given two hospital instances, if their values for at least one of the properties \texttt{Address} and \texttt{Contact Details} are the same, they are treated as equivalent. Given two physician instances, if their values for the property \texttt{workAt} are equivalent, and the values for the properties \texttt{Name} and \texttt{Title} are the same respectively, these two instances are determined as equivalent. Equivalent instances are fused as one instance in AsdKB.

\subsection{Other Knowledge}
In this part, we extract factual knowledge on the instances of intervention methods, and the information of China administrative divisions. Instances of intervention methods are obtained from The National Clearinghouse on Autism Evidence and Practice\footnote{\url{https://ncaep.fpg.unc.edu/}} (NCAEP), and such instances are all evidence-based practices, including ``Discrete Trial Training'', ``Social Skills Training'', ``Peer-Based Instruction and Intervention'', and etc. For each instance of intervention methods, we extract the values of properties \texttt{Label} (instance name) and \texttt{Introduction} (a brief description on the instance information). English string values are translated to Chinese by Google Translate, and we also conduct careful proofreading.

With expert knowledge introduced in Section~\ref{subsec:ek}, a potential application is to seek expertise help from physicians to diagnosis. In order to find professional physicians in the target districts, cities, and provinces, we extract instances of China administrative divisions from National Bureau of Statistics\footnote{\url{http://www.stats.gov.cn/}}. The extracted instances are specific districts, cities, and provinces, and we also build \texttt{locateAt} relationships among them. To link each hospital to the corresponding administrative divisions, we first use Amap (a leading provider of digital map in China) API\footnote{\url{https://github.com/amapapi}} to get the latitude and longitude of each hospital by inputting the value of property \texttt{address}. With the information of latitudes and longitudes, Amap API can return the corresponding districts, cities, and provinces of hospitals. Besides, we record the \texttt{Population} of each instance of China administrative divisions, which could help regional analysis on ASD.

\subsection{Quality of AsdKB}\label{subsec:qoa}
AsdKB contains 6,166 entities (including conceptual entities, i.e., classes, and individual entities, i.e., instances) and 69,290 triples in total. All class URIs in the namespace \textbf{\url{http://w3id.org/asdkb/ontology/class/}} and instance URIs in the namespace \textbf{\url{http://w3id.org/asdkb/instance/}} are dereferenceable. To evaluate the quality of AsdKB, we design two evaluation methods: accuracy evaluation, and task evaluation.

\spara{Accuracy Evaluation.}
There is no ground truth available, and it is impossible to evaluate all triples manually. Therefore, we apply a random evaluation strategy. We first randomly select 100 entities distributed across classes and instances, and obtain 732 triples. These samples can reflect the distribution of triples in the entire knowledge base. We then conduct manual labeling to evaluate the accuracy of the samples. The accuracy of the entire AsdKB is estimated by evaluating the accuracy of the samples.

Five graduate students participate in the labeling process. We provide three choices, which are \textit{correct}, \textit{incorrect}, and \textit{unknown} to label each sample. After each student label all the samples, we calculate the average accuracy. Finally, similar to YAGO~\cite{hoffart2013yago2}, Zhishi.me~\cite{wu2020knowledge}, and Linked Open Schema~\cite{wu2018building}, we use the Wilson interval~\cite{brown2001interval} when $\alpha = 5\%$ to extend our findings on the subset to the entire knowledge base. The Wilson interval is a binomial proportion confidence interval calculated from the results of a series of Bernoulli trials, and $\alpha$ is the significance level. For the randomly selected 732 triples, the average \textit{correct} votes is 712, so the accuracy is 97.02\% ± 1.21\%, and it demonstrates the high quality of AsdKB.

\spara{Task Evaluation.}
Besides the accuracy of the triples in AsdKB, we try to evaluate the effectiveness of AsdKB in answering real-world ASD relevant questions. Thus, we collect 100 frequently asked questions (e.g., ``\begin{CJK*}{UTF8}{gbsn}孤独症都有哪些临床表现？\end{CJK*} (What are the clinical symptoms of autism?)'' and ``\begin{CJK*}{UTF8}{gbsn}哪些干预方法是有效的？\end{CJK*} (Which interventions are effective?)'') on ASD from Chinese healthcare websites The Good Doctor and Family Doctor (introduced in Section~\ref{subsec:ek}), which are also the data sources of the expert knowledge in AsdKB. We store AsdKB in a graph database Neo4j~\cite{webber2012programmatic}, and also invite five graduate students to manually write Cypher (Neo4's graph query language) queries for the collected questions so as to check whether the returned query results can answer the questions. According to the above evaluation, AsdKB can answer 81 questions, i.e., the coverage reaches to 81\%, which reflects the practicality of AsdKB.

\section{Application of AsdKB}\label{aoa}
To illustrate the potential application of AsdKB, this section describes the implementation of a prototype system\footnote{\url{http://asdkb.org.cn/}} for the early screening and diagnosis of ASD based on AsdKB. This system has three main applications, including question answering, auxiliary diagnosis, and expert recommendation. Users of this system are parents, teachers, and caregivers. 

\subsection{Question Answering}
We implement a natural language question answering (QA) system based on AsdKB, and expect that the QA system can answer various common-sense and factual questions on ASD. As mentioned in Section~\ref{subsec:qoa}, AsdKB is stored in Neo4j, so we aim to translate each natural language question to a Cypher query, in order to query the graph database to return the answer. We use two strategies to design the QA system. The first one is to manually write common ASD relevant natural language query patterns (i.e., regular expressions) according to AsdKB ontology and the corresponding Cypher query templates. If a user query matches one of our patterns, then we construct and execute the corresponding Cypher query based on the pre-defined Cypher query template to get the answer. If the user query does not match our patterns, we use the second strategy, which applies the idea of the method for translating natural language questions to formal queries with semantic query graph modeling~\cite{zou2014natural} to generating the Cypher query.

\begin{figure}[t]
  \centering\includegraphics[width=1\textwidth]{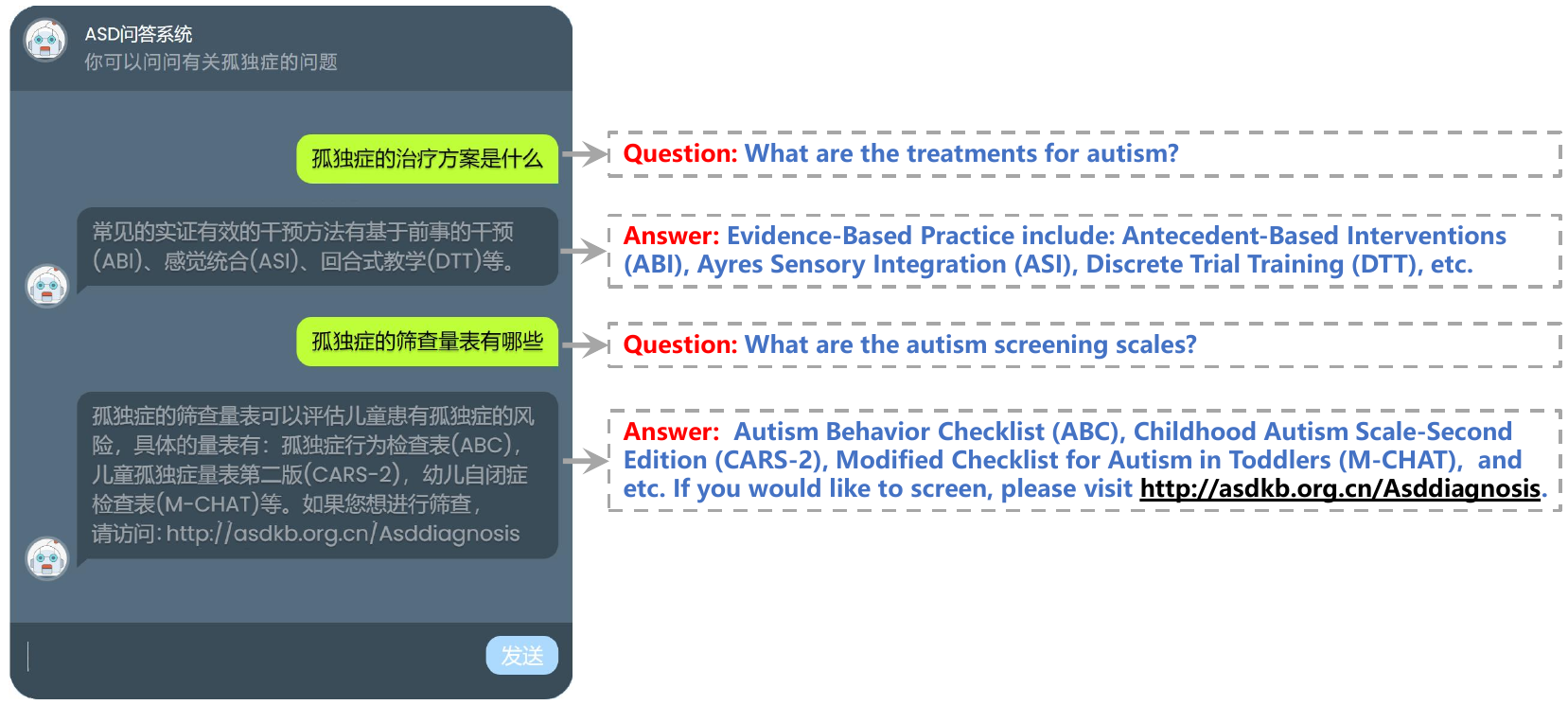}
\caption{The interface of the QA system.}
\label{fig:qas}
\end{figure}

Figure~\ref{fig:qas} shows the interface of our QA system. We also detect the intention of each question to check whether the user would like to further fill in screening scales. If so, the system will directly give the link of auxiliary diagnosis to help choose screening scales (see Figure~\ref{fig:qas}). The intention identification is modeled as a task of binary classification, where we use BERT to encode questions in the labeled data, and then train a SVM~\cite{hearst1998support} classifier to predict whether users are willing to conduct screening or not.

\subsection{Auxiliary Diagnosis}
We have developed an auxiliary diagnosis system based on AsdKB. This system provides users with screening scales to assess the risk of being ASD. As long as the screening result of a screening scale shows a risk, the system will prompt the user to seek professional medical evaluation and recommend experts using our expert recommendation system (will be introduced in Section~\ref{subsec:er}).

\begin{figure}[t]
  \centering\includegraphics[width=1\textwidth]{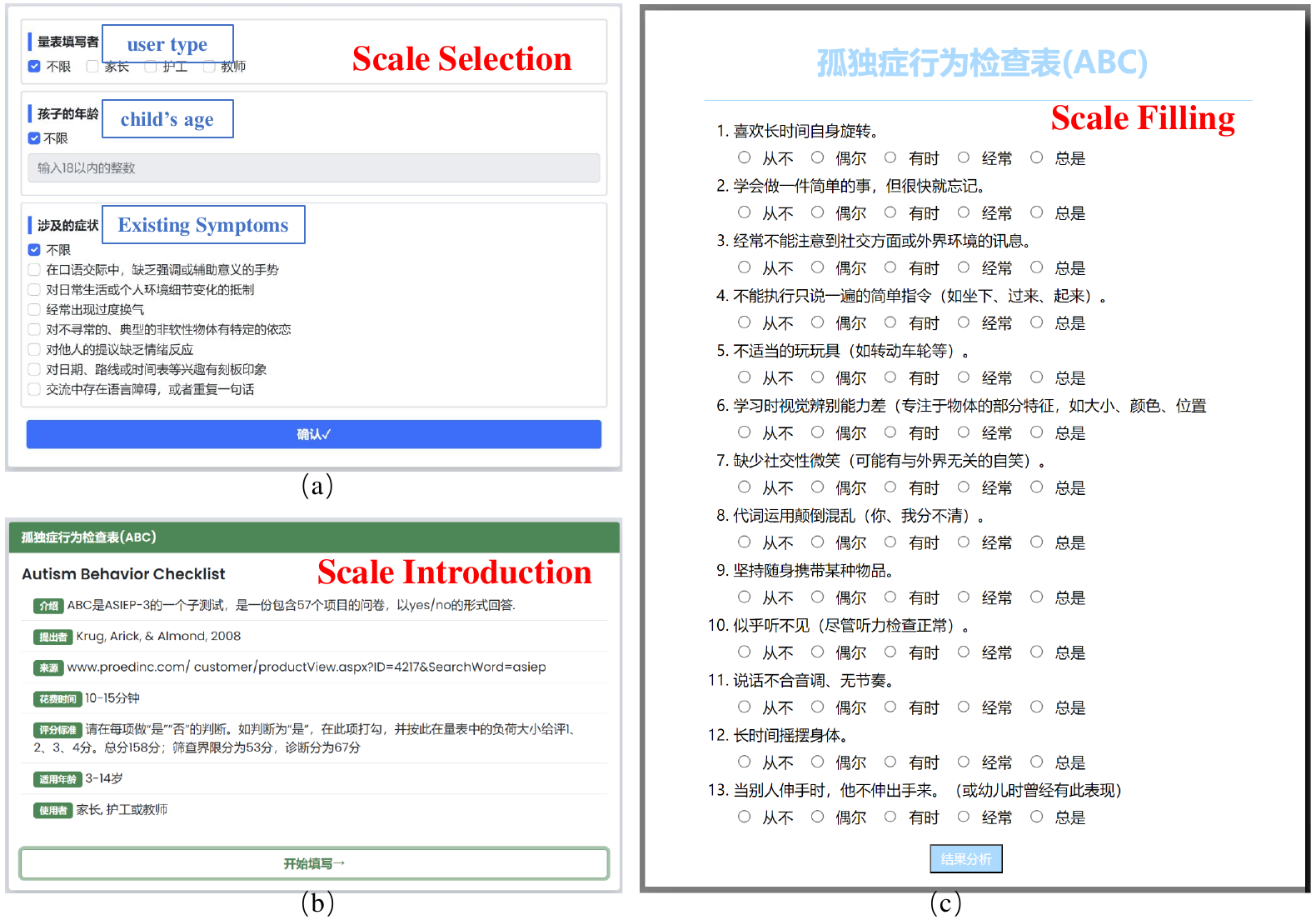}
\caption{An illustration of the auxiliary diagnosis system.}
\label{fig:ad}
\end{figure}

As shown in Figure~\ref{fig:ad}(a), before filling the screening scales, users can select appropriate screening conditions based on their situations, such as the child's age and existing symptoms, and the system will return the corresponding screening scales with a brief introduction (see Figure~\ref{fig:ad}(b)). Figure~\ref{fig:ad}(c) shows the questions and options when filling in the ABC screening scale. After completing a screening scale, the system will give the screening result (i.e., risky or not) based on the total score of all options and the screening boundary.

When users are filling in screening scales, they can check what specific symptoms the current question investigates to better understand the question, so as to help make a choice more precisely. Besides, after completing screening scales, this system can also analyze which option matches some diagnostic standard, to provide explanations of the screening results. More details have already been introduced in Section~\ref{subsec:dk} and Figure~\ref{fig:dkEXP}.

\subsection{Expert Recommendation}\label{subsec:er}
\begin{figure}
  \centering
  \includegraphics[width=1\textwidth]{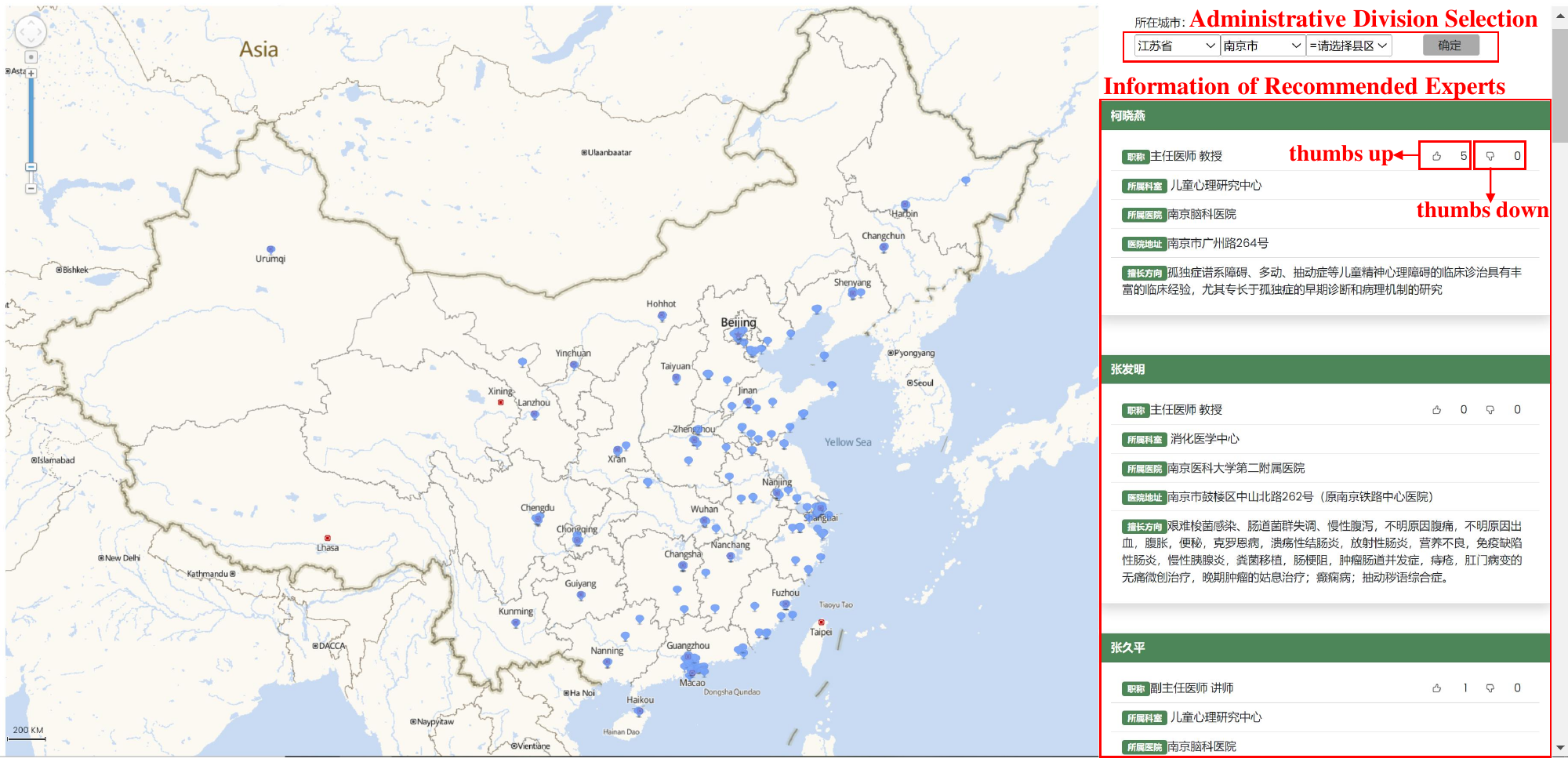}
\caption{An illustration of the expert recommendation system.}
\label{fig:er}
\end{figure}

If our auxiliary diagnosis system reports the risk of being ASD, users may have requirements to find experts on diagnosing ASD in the target administrative divisions. Thus, we design an expert recommendation system with facet search on AsdKB. Users can choose the target province, city and district by selecting a checkbox or directly clicking their locations on the map (see Figure~\ref{fig:er}). The recommendation result is a list of professional physicians with their names, titles, hospital departments, hospitals, hospital addresses, and specialties.

The recommendation has two steps: candidate physician generation and candidate physician ranking.
In candidate physician generation, we use the location information of hospitals in AsdKB to match user selected administrative divisions, and the physicians in AsdKB working at such hospitals are candidates. Note that if no candidate physician returns, we will consider more hospitals in surrounding administrative divisions by distance calculation with latitudes and longitudes. In the candidate physician ranking, three aspects are taken into consideration. Firstly, the higher the title, the higher the ranking. Secondly, the higher the hospital level, the higher the ranking. Finally, the higher the number of thumbs up minus the number of thumbs down (Figure~\ref{fig:er} gives an example), the higher the ranking.

\section{Related Work}\label{rw}
Tu et al.~\cite{tu2008using} first proposed an autism ontology with domain terms and relationships relevant to autism phenotypes. The main target is to enable user queries and inferences about such phenotypes using data in the NDAR repository, but it does not include DSM criteria, so it does not support diagnosis of ASD. McCray et al.~\cite{mccray2014modeling} also developed an ASD-phenotype ontology assessing and comparing different ASD diagnostic instruments, but it also does not include DSM-IV or DSM-5 criteria phenotypes. ADAR~\cite{mugzach2015ontology} extends an ontology proposed by Tu et al~\cite{tu2008using}. with additional SWRL rules to infer phenotypes from ADI-R~\cite{lord1994autism} items, and it covers various symptoms and features of DSM IV and DSM-5 diagnostic criteria, such as difficulties with social interaction, language and communication issues, and stereotyped and repetitive behaviors. However, many fine-grained classes are actually instances in the generic sense. 

The most recent work is AutismOnt~\cite{hassan2022autismont}, an ontology for autism diagnosis and treatment, which covers various covers autism research directions. AutismOnt includes the classes: Diagnosis, Risk Factors, Treatments, Strength and Weakness, Services, Lifespan Issues, Profile, and Family Relationships. However, although the authors claim that AutismOnt is available in the NCBO BioPortal, it cannot be found in the repository. 
 
Some large-scale medical knowledge bases also contain ASD knowledge. For example, SNOMED CT~\cite{donnelly2006snomed} contains a large-scale number of medical terms, and the disease classes in AsdKB also comes from SNOMED CT, but it does not cover other kinds of knowledge, such as diagnostic knowledge and expert knowledge. Yuan et al.~\cite{yuan2020constructing} proposed a method for constructing knowledge graphs with minimal supervision based on unstructured biomedical domain-specific contexts. They collected 24,687 abstracts of articles related to ASD from PubMed\footnote{\url{https://pubmed.ncbi.nlm.nih.gov/}}, and constructed a knowledge graph on ASD. However, they did not design the ontology and the knowledge graph is not publicly available. CMeKG~\cite{byambasuren2019preliminary} is a Chinese medical knowledge graph developed using natural language processing and text mining techniques from a large amount of medical text data. CMeKG mistakenly uses drugs as the treatment for ASD, but drugs are only used to alleviate the complications of ASD in fact.

Compared with all existing works, AsdKB is the first publicly available Chinese knowledge base on ASD, and it contains both ontological and factual knowledge about diseases, diagnosis, experts, and others. AsdKB has been applied in developing applications of the early screening and diagnosis of ASD.

\section{Conclusions and Future Work}\label{cfw}
We develop and publish a Chinese knowledge base on ASD called AsdKB by extracting and integrating knowledge from various data sources with different formats. To the best of our knowledge, AsdKB is the 
most comprehensive ASD knowledge base on the Web, and it supports the different applications on the early screening and diagnosis of ASD, such as question answering, auxiliary diagnosis, and expert recommendation. However, there are still some limitations to our work that we plan to address in the future.

\spara{Quality of AsdKB.} During our preliminary evaluations of AsdKB, we discovered that the entities contained within the knowledge base are of high quality. However, errors do exist during the automatic extraction process. These errors stem from a variety of factors such as the quality of the original data sources, differences in data formats, and our integration methods. To address this issue, we plan to introduce crowd-sourcing techniques to fix the existing errors in AsdKB and study automatic error detection methods to ensure the accuracy of knowledge in the process of knowledge update.

\spara{Applications of AsdKB.} We have explored various applications for AsdKB, including QA, auxiliary diagnosis, and expert recommendation. The integrated prototype system has demonstrated the potential for AsdKB to play a critical role in early ASD screening and diagnosis. To further improve the accuracy of QA and auxiliary diagnosis, we will incorporate data-driven machine learning models on more user log data in our prototype system. In addition to this, we plan to analyze electronic medical records if possible using AsdKB to assist physicians in ASD diagnosis. By analyzing medical histories, symptoms, and other relevant information using AsdKB, physicians can make more accurate diagnosis and give appropriate and personalised treatment suggestions to the people with ASD.

\subsubsection{Acknowledgements}
This work is supported by the NSFC (Grant No. 62006040, 62072149), the Project for the Doctor of Entrepreneurship and Innovation in Jiangsu Province (Grant No. JSSCBS20210126), the Fundamental Research Funds for the Central Universities, and ZhiShan Young Scholar Program of Southeast University.
\bibliographystyle{splncs04}
\bibliography{reference}
\end{document}